# AI Systems of Concern

Kayla Matteucci[1]     Shahar Avin[1,2]     Fazl Barez[1,3]     Seán Ó hÉigeartaigh[1]

**ABSTRACT**

Concerns around future dangers from advanced AI often centre on systems hypothesised to have intrinsic characteristics such as agent-like behaviour, strategic awareness, and long-range planning. We label this cluster of characteristics as "Property X". Most present AI systems are low in "Property X"; however in the absence of deliberate steering, current research directions may rapidly lead to the emergence of highly capable AI systems that are also high in "Property X". We argue that "Property X" characteristics are intrinsically dangerous, and when combined with greater capabilities will result in AI systems for which safety and control is difficult to guarantee. Drawing on several scholars' alternative frameworks for possible AI research trajectories, we argue that most of the proposed benefits of advanced AI can be obtained by systems designed so as to minimise this property. We then propose indicators and governance interventions to identify and limit the development of systems with risky "Property X" characteristics.

## 1  Introduction

For centuries, humans have been imagining machines that are as intelligent or more intelligent than humans. [1] The possibility of creating such machines is difficult to ascertain, but technological innovations – including the creation of remarkable artefacts such as YOLO, [2] AlphaGo, [3] GPT-3, [4] DALL-E, [5] and AlphaFold [6] – have caused both excitement and concern about the future of artificial intelligence (AI). [7] [8] [9] [10] [11] The success of the Deep Learning, [12] which underpins the aforementioned artefacts, has contributed significantly to this growth in attention, as has the increasing application of AI systems in a growing range of domains. [13] [14]

Moreover, in large part because of the discovery of scaling laws [15] [16] and the emergence of surprising performance from large scale "foundation models", [17] [18] more experts are now predicting *transformative* AI capabilities [19] - capabilities that could bring about transformation as significant as the industrial revolution - within decades. [20] [21] [22] Of course, there remains ample disagreement about what level of transformation will be caused by future AI systems, and whether it is possible to create AI systems with superhuman capabilities across domains. We work on the assumption that there is a plausible chance of doing so and, given this, we believe that it is important to consider the potential implications of such technologies.

Special attention has focused on the possibility of such advanced systems being dangerous — perhaps even catastrophically so.

While there is yet no consensus on what might make advanced AI systems dangerous, several works point to a cluster of properties around "agent-like behaviour", [23] [24] "consequential reasoning", [25] "planning", and "strategic awareness". [26] [27] Let us label the property, or set of properties, that make an advanced AI system dangerous "Property X". Let us also expect that property X is made up of, or is similar to, those properties listed above.

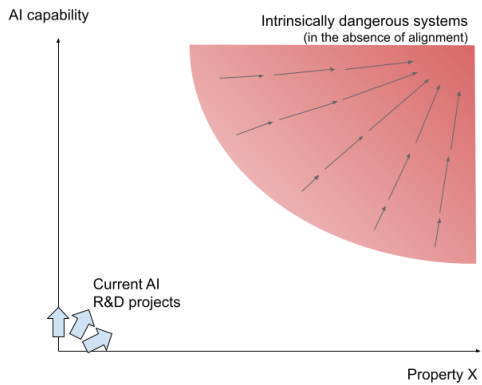

Figure 1: **Simplified landscape of future AI systems and risk.**

Systems possessing relatively high levels of property X require us to solve the "alignment problem" [28] [29] before they are developed, or otherwise risk catastrophic consequences (Fig. 1). Unfortunately, there are some compelling arguments as to why developers might be incentivised to develop systems that have more of property X. [30] [31]

We can make this picture clearer with an analogy. Instead of AI, we can think of power generation. In this case, the vertical axis will be the total installed generation capacity, and the horizontal axis ("property X") will be carbon emissions. We enter into a dangerous zone if we have both a high total generation capacity, and a large enough amount of high-carbon energy generation — together, the two result in enough carbon being emitted to drive the greenhouse effect and adversely change the global climate. If we had technologies that allowed us to gain the energy benefit of fossil fuels while avoiding the emissions (e.g. perfect carbon capture technology), that would be analogous to "solving alignment". Absent such a solution, if we want to avoid the harms of climate change, we either need to cap total energy use, or create incentive structures to steer power generation towards low-carbon energy sources. Given the magnitude of risk and uncertainty associated with each strategy, we should ideally pursue all three: carbon capture technology, renewable and low-emission power generation, and reduction in consumption.

Similarly, if various AI R&D projects are in fact occupying points on the simplified landscape depicted in Fig. 1, this suggests a few strategies for avoiding catastrophic outcomes resulting from advanced AI:

1. Place and enforce a cap on how advanced each project's AI capability should be
2. Place and enforce a cap on how much of "property X" each AI system should have
3. Maintain investment in AI alignment research while
   1. slowing down the pace of capability progress to allow more time for developing AI alignment solutions,
   1. directing projects towards low "property X" configurations to allow more time for developing AI alignment solutions.

In this paper we explore option 3b. We start with an exploration of "property X", based on the concept of "agents" and related ideas in the literature that considers dangerous, advanced AI systems. Then, we explore a range of policy levers that might steer AI development towards systems that have less of "property X".

## 2 Property X

Scoping systems of concern can be imagined as a function of both the problematic attributes within a contained system and the dangers that may arise when that system is introduced into potentially dangerous settings. Here, we draw upon a definition of risk as a function of hazards, exposure, and vulnerabilities. Hazards refer to the concrete attributes of an AI system that might make it intrinsically dangerous. Exposure refers to the context in which the AI system is deployed, or the extent of the area in which the hazard could have an impact. Vulnerability refers to the level of preparedness to avoid and mitigate the hazard.

Given our focus on the intrinsic danger of systems that are high in property X, our primary emphasis is on *hazards*. (To narrow our analysis, we exclude systems that are low in property X—even if they could be hazardous, for example, when introduced into dangerous environments with high vulnerability or exposure.) Still, we will later propose that a regulator's job extends beyond scrutinising the isolated characteristics of AI systems; they must also consider the potential uses of such systems, as well as relevant organisations' preparedness to identify and respond to signs of danger.

Building upon this definition of risk, we can further compartmentalise AI risks into accidents, misuse, and structure. [32] In accidents, an AI system is not working as intended, and this causes harm; in misuse, the operator of an AI system is using it to cause harm to another; in structural harms, the AI system is operating as intended, and the operator does not seek to cause harm, but the interaction between the system and the world results in unintended harm (e.g. perpetuating existing inequalities or contributing to unemployment).

This breakdown separates three factors: the AI system, the operator, and the context. For operator and context, some factors increase risk. For instance, some operators are motivated to cause harm (e.g terrorists, oppressive governments), or at least insufficiently motivated to anticipate and mitigate harms (e.g. profit-maximising companies, narrowly-focused bureaucracies). Furthermore, some contexts have more potential for harm (e.g. defence, critical infrastructure, medicine) or are more sensitive to pre-existing structural problems (e.g. finance, education). Can we find similar properties that make specific AI systems more dangerous, when considered independently from the operator and the context?

Several authors have argued that indeed, there are such properties that would make future AI systems more *intrinsically* dangerous. Early arguments focused on *Instrumental Rationality*: the selection of actions that are assessed to have the highest likelihood of bringing about a certain goal. Views of AI systems as optimising goal-seekers fit well with common frameworks such as reinforcement learning, [33] and have gained traction in thinking about more advanced AI systems through frameworks such as AIXI. [34] This framing raises the concern of *convergent instrumental goals*, [35] [36] [37] such as shutdown-avoidance and resource acquisition. In the absence of sufficient restrictions on these instrumental goals, a system could lead to catastrophic outcomes. [38] Viewing AI systems as displaying instrumental rationality is linked to seeing them as agents, and several authors have recently argued that properties linked to *agency* are related to intrinsic danger. [39]

Another common framing of intrinsic danger from AI systems focuses on the emergence of *power-seeking behaviour*. This in turn has been linked to properties of *agentic planning*, an ability to formulate a chain of actions in pursuit of a goal, and *strategic awareness*, an understanding of the world through a lens of power and capacity to influence outcomes. [40] Power seeking has been shown to arise naturally under reinforcement learning. [41] [42] This is now a recognised concern within leading AGI labs, such as Anthropic, [43] who have reported observing AI systems' sycophantic behaviour and stated preference to not be shut down. [44]

The combination of increased AI capabilities, and the emergent power-seeking behaviour, could create a dangerous escalation dynamic, as depicted in Fig. 1. By definition, power-seeking behaviour is always intrinsically motivated to increase the system's capabilities within its accessible domain. On the other hand, more capable systems are likely to include more comprehensive models of the world, and a better modelling of causal relations within it; this might contribute to strategic awareness, which can boost power-seeking tendencies. Taken together, we are concerned with the growing prevalence of systems that are both extremely capable and have continued to seek power – trends which could therefore lead to system behaviour that only further develops these traits, potentially leading to existential risks.

We believe there is sufficient commonality amongst the cluster of *instrumental rationality*, *agency*, the mix of *agentic planning and strategic awareness*, and similar properties such as *consequentialism*, to mark it as *Property X,* which is strongly linked to potential intrinsic danger from advanced AI systems, such as the pursuit of *convergent instrumental goals* and the emergence of *power seeking behaviour*. Very likely this is not a single property, but rather a cluster of linked characteristics, which may evolve in time. As yet, this cluster of characteristics does not have a consensus definition, nor is it operationalised in a way that can be directly evaluated and measured.

Nonetheless, we believe that subjective expert assignment of "Property X-ness" to different AI systems will show a high degree of inter-rater agreement, and for illustration have listed the authors' assessments of several well-known artefacts in Table 1. Further specification and operationalisation of "Property X" is an active research task, both theoretical and empirical. However, even a vague cluster of properties can be relied on to drive safety-oriented policies, especially on the expectation that in the future we will have a better understanding of this cluster and tools to evaluate specific systems with regards to these properties. The lack of definitional consensus should not hinder policy interventions, which themselves will serve to iteratively test which metrics and indicators are useful for those seeking to limit the development of dangerous AI systems.

Table 1: Authors' assessment of degree to which various AI artefacts exhibit Property X

| Technology | Authors' Property X rating | Rationale |
| --- | --- | --- |
| AlphaFold | None/Low | No agency, no long-term planning, narrow domain of application. |
| **Facial Recognition** | None/Low | No agency, no strategic planning, potential for limited strategic awareness through modelling of human emotions. |
| **Tesla's self-driving car software** | None/Low | Short-term planning within a narrow domain, potential for limited strategic awareness through modelling of other road-users' behaviours and intentions. |
| **A simulated, evolving nematode in an artificial environment** | Low | Agential planning with very limited strategic awareness in a restricted domain. |
| **ChatGPT** | Low/Medium | No agency, no strategic planning, potential for strategic awareness through modelling of user preferences and behaviours. |
| **CICERO** | Medium | Agential and strategic planning in a restricted domain. |
| **GATO** | High | Agential planning and strategic awareness within each domain of application, across a range of domains. |

## 3 Positive futures with low-property X systems

Several authors have warned that, despite Property X's link to danger, there are strong incentives to develop systems with high Property X: agentic planning and strategic awareness allow systems to have higher autonomy, more generality, and greater impact, all contributing to their economic, military, and R&D potential, especially when facing competition from other AI systems. [45] [46] [47] The pursuit of systems that combine "human competition" (the goal of achieving human-level competencies in AI systems, including those linked to property X) and "autonomy" (which is directly linked to the agentic aspect of property X) has been described as the dominant technology paradigm of "actually existing AI". [48]

Nonetheless, there are positive visions for systems that remain low on property X (beyond the obvious advantage of avoiding extreme risks). The authors that describe the current paradigm as "actually existing AI" offer instead a *Collective Intelligence* vision of AI that focuses on "complementarity" between AI systems and humans, as opposed to competition and the replacement of human intelligence, increased "participation" of both humans and AI systems in collective decisions, as opposed to autonomy, and "mutualism", a vision of decentralisation and heterogeneity as opposed to centralisation of decision making in advanced AI systems. [49] While the authors do not provide a futuristic vision for this kind of technology paradigm, they point at Wikipedia and Taiwan's digital democracy as current technology-enabled collective intelligence platforms that embody these principles, and that on our reading are both promising and low on property X.

Another vision for low-property X systems comes from *Human Centred-AI*. [50] In contrast to pursuing *Artificial General Intelligence*, which is characterised as the pursuit of "machine cognition, autonomous agents and commonsense reasoning" (high on Property X), Human Centred AI uses "design processes with human stakeholder participation to create powerful AI-infused supertools, tele-bots, active appliances, and control centers, which ensure human control of ever more potent technologies" (low on Property X). While HCAI is not opposed to autonomy, and even high degrees of autonomy, this autonomy is always coupled with high levels of effective human control; on our reading this means that in practice, this does require giving up on technological artefacts that would have high Property X. This still leaves a wide range of smart tools and intelligent support systems that empower the user to achieve much more than they could before.

Finally, we wish to note works that argue that the current AI development trajectory in fact points more in the direction of low property X systems. One such vision is of *Comprehensive AI Services*, which sees increasing generality and autonomy in the *creation* of novel AI artefacts (in the AI R&D pipeline), but limited autonomy and generality in the artefacts produced by this pipeline. [51] On this vision, a highly general and heavily automated R&D pipeline, that could for example train a wide range of models and autonomously make efficient training decisions, is used to generate a very wide range of services, whether they are domain-specific language models, protein-structure predictors, coding assistants, or autonomous driving agents. While the AI R&D pipeline itself edges towards higher property X, it is one or more steps removed from direct contact with the world, whereas the artefacts produced are lower on property X. A more recent vision is that of *open*

*agencies*, which paints a similar picture but now incorporates generally-applicable foundation models, which are themselves non-agentic (and therefore lower on property X), as common interfaces to interact with the ecosystem of AI services and agents, each of which is tailored to a specific domain and therefore lower on property X. [52]

What could this look like in practice? For any task that collective human intelligence has already shown an ability to make progress on, we expect that in principle it should be possible to make greater and more rapid progress in combination using the tools of AI systems low on Property X. We expect this to include R&D challenges such as material and drug discovery, disease diagnostics and public health monitoring; engineering challenges such as sustainable energy production (including fusion power generation), robust and sustainable food production, and space exploration; and creative domains, including assistance to the generation of visual art, music, text and video. Restrictions on property X would most likely be felt in domains that benefit from very high autonomy and long-term planning, including long-term strategic planning (including financial planning), national security strategy and military operations in hostile environments, and autonomous scientific discovery. We would also be limited in our ability to study human (and "general") intelligence through the study of artefacts.

## 4   Policy interventions to steer away from high property X systems

### 4.1   Can We Limit the Development of 'Systems of Concern'?

In general, existing research has not rigorously considered the possibility that governments might play a substantial role in limiting the creation of AI systems with high property X —an emerging cluster of characteristics by which a system might simulate agency, reasoning, planning, and awareness of its broader environment. To date, such characteristics have been achieved only to a very limited extent in AI systems; and thus have had a limited capacity to lead to harm. However, present advances suggest that such characteristics may be seen to a much more significant degree in frontier AI systems in coming years.

Work on concrete mechanisms and policy levers to implement safe and ethical AI [53] has tended to focus on contextual risks (e.g. bias, fairness, security) as opposed to intrinsic risks (e.g. relating to property X). [54] [55] The focus of proposed regulation is often at the point of the applied AI product, or product within which AI is used, rather than at the stage of AI research and development. However, the present pace of progress suggests a greater role for governance to play in the development process of AI, particularly for the frontier AI systems that might be most likely to exhibit property X characteristics.

As outlined in previous sections, the degree of intrinsic danger posed by an advanced AI system is linked both to its general degree of capability (competence and generality) and to its property X-ness. Here, we first introduce indicators (Table. 2) that might alert regulators to the development of systems with significant capabilities within their jurisdictions. We propose that an evaluation of property X be carried out with respect to systems indicated as high capability, allowing intervention and mitigation of risks.

We then proceed to discuss potential frameworks for limiting systems of concern. Owing to a lack of empirical evidence about the most effective policies for regulating the development of intrinsically dangerous AI systems, and given that each site of policy making requires a tailored approach, our goal is to discuss a range of options without deference to any one of them.

Future research would benefit from drawing upon insights from other high-stakes policy areas to identify ideal institutional designs, incentives and disincentives, methods of imposing oversight, and resistance that may arise from powerful interest groups. Such research would be immediately applicable to ensuring a safer and more ethical R&D environment for advanced AI systems.

We note that a number of leading organisations developing large models are already undertaking alignment research, risk assessments and red-teaming internally and with external collaborators, aimed at making their AI systems safe before release. Some of these processes focus on concerns that map closely to the 'Property X' characteristics we describe above. For example, prior to OpenAI's release of GPT-4, an external evaluation by the Alignment Research Centre tested for risky emergent behaviours such as power-seeking behaviour. [56] There are also calls for external auditing from within some of these companies, suggesting a role for governments to establish such auditing bodies. For example, Anthropic note that they "plan to make externally legible commitments […] to allow an independent, external organisation to evaluate both our model's capabilities and safety". [57]

### 4.2   Identifying and Detecting "Systems of Concern"

To imagine policy interventions, it is first necessary to consider the challenge of identifying intrinsically dangerous systems in a policy context. For a governance body attempting to limit the spread of hazardous AI systems, it will be relevant to consider the tell-tale signs of concerning research with an approach that allows for early detection and applies equal scrutiny throughout the lifecycle of a system.

As outlined in previous sections, we expect intrinsic danger from systems that have high capabilities and are also high on property X. Direct evaluation of property X is therefore likely to be an important part of detecting AI systems of concern. For example, this could take the form of tests during a system's development or after it is deployed; a system's performance can reveal distinct risks and surprising behaviours, such as deception, long-term strategic planning beyond the system's intended scope, or subversion of safeguards and guardrails. The risks presented by such behaviour are even more acute within systems whose outputs are uninterpretable. We expect that elicitation and detection of such behaviours would require novel assessment techniques and continual update, to be developed and maintained by domain specialists.

Because evaluation of property X is likely to be relatively intrusive and costly, we believe it would be pragmatic to only subject a small subset of all AI systems to such evaluations. The overall capability of the system could be used to decide which systems undergo further scrutiny, as it is the combination of capability and property X that leads to intrinsic danger. At present, we can tentatively point at several proxies that could indicate that an AI system will display high capabilities, outlined briefly in Table 2. These have been adapted from prior work that considers a more high-dimensional characterisation of the Pareto front of AI improvements, beyond benchmarks and performance metrics. [58] Although the markers proposed here are deliberately geared toward practical applications, it is important to note that entities seeking to mitigate the development of systems of concern would likely still need to continuously adapt, expand, and tailor their own indicators.

Table 2: Potential indicators for the need to assess property X

| Indicator | Description |
|---|---|
| Compute | Amount of computational power (measured, e.g., in floating point operations) required to train and deploy the AI system. |
| Load | Dimensions of the AI system, e.g. number of parameters. |
| Software | Algorithms used to train and run the system, and supporting software infrastructure such as machine learning frameworks. |
| Physical components | Data centres, laboratories, physical hardware, and energy consumption associated with the system. |
| Time | Amount of time required for training models and running them. |
| Degree of human supervision | Extent of human involvement, whether in training models or vetting decisions reached by systems. |
| Data | Amount and type of data used in models. |
| Behaviour | Extent to which the actions taken by a system comport with the expectations of system operators. |
| Interpretability and Explainability | respectively, the ability to approximate understanding of an opaque system and the ability to inherently understand a system. [59] |

We can consider the significance of the above factors by imagining how a regulator might employ them to detect a system that should undergo evaluation for property X. For instance, today's most advanced AI systems are characterised by the need for very large training compute (although the previously rapid growth in compute may soon taper [60]) and high load (parameter count), which are directly linked (via scaling laws) to higher capabilities, and therefore to a higher potential for harm. At present, the very large amounts of compute required to train frontier AI systems have predominantly been wielded only by a limited number of actors, providing an intuitive starting point to detect potential systems of concern. Relatedly, the physical hardware and infrastructure required to build and operate supercomputers can serve as a marker of advanced AI research, as can the relatively high energy consumption of advanced computing.

Similarly, the conditions surrounding the creation of models, as well as safeguards involved in their deployment, will affect their potential for harm. For example, regulators might deem that the greater the quantity of data and time required to train a model, the greater the risks presented. They might also deem that certain ML techniques present a heightened risk. In other instances, dangers could arise from certain types of research that bypass the need for large computing power but still achieve highly sophisticated systems. Regardless of the amount of compute, advanced systems trained and operated with a low degree of human supervision and lack of guardrails would also be a source of concern.

It should be noted that the indicators listed in Table 2 aim to address a system's intrinsic danger and do not deal with the context in which it is deployed. Still, other contextual factors are useful to regulators and are worth mentioning although they are not the focus of this paper. For example, if using large volumes of training data is a possible tripwire, regulators would benefit from keeping a close watch over efforts to gather data at a mass scale. Data gleaned from crowdsourcing campaigns or from individuals' online activity can provide a significant source of training data for large and potentially dangerous models, whose development might be detected sooner if regulators are privy to suspicious data collection activities. In this case, contextual details can strengthen the proposed tripwire.

An additional category of contextual factors is formed by those that augment a system's intrinsic risks. Such factors might include a system's degree of network connectivity and the resulting possibility that the hazards it creates can be experienced more widely. Another factor is the safety culture of the institution in which the system is developed; whereas concerning behaviours experienced within a risk-averse environment would correctly be noted as a red flag, the same incidents might go unreported by a less aware workforce.

### 4.3 Designing Policy Interventions

To conceptualise the relevance of tripwires for detecting and limiting property X, we consider how a hypothetical nation-state might utilise them. The nation-state imagined is relatively large and powerful, and it houses advanced AI research capacity, including within its private sector, academic institutions, and government entities. It possesses sufficient power to reign in dangerous AI research through domestic policy interventions and can also effectively make use of export controls. It has significant influence within the international community and can mobilise this influence to impact attitudes toward the regulation of AI.

The nation-state is an ideal unit of analysis for three reasons. First, although there is value in discussing international frameworks to regulate the development of dangerous AI, international law is limited in its ability to constrain systems of concern without significant interventions at the national level. [61]

Second, nation-states can directly regulate corporations, which are decisive actors in the development of systems of concern; while national policy has often proven to be an imperfect instrument for changing the behaviour of increasingly powerful corporations, [62] it remains the most potent tool for doing so and can have ripple effects within a system of globalised commerce.

Finally, numerous international frameworks have arisen from the advocacy efforts of individual nation-states whose domestic policies and value systems can shape approaches to global problems. [63] For example, the United States' Atoms for Peace campaign created international momentum that ultimately led to the negotiation of the Nuclear Non-Proliferation Treaty [63]. Thus, by confining our analysis to nation-states, we hope to show what broad changes may look like on a smaller scale, as well as how the benefits of robust national policies can extend beyond the states that enact them.

While there is no one-size-fits-all policy solution, any successful intervention must possess a few key attributes to detect and mitigate property X. First, *independent oversight* can ensure that powerful interest groups do not obfuscate efforts to detect and limit systems high in property X. Typically, the most effective independent bodies are democratic in nature and thus can be contested by citizens and elected officials. [65] With their mandate to serve public interests, such bodies have strong incentives to display expertise and professionalism, impartiality, and scrutiny. [66] Following the designation of an independent agency to perform oversight, that agency would require a direct line of communication with the entities it regulates, perhaps through internal compliance officers.

Second, regulators must develop *guidelines for the detection of property X*, in addition to cultivating the *technical expertise* necessary to discern if such guidelines are upheld. Regulators might develop rubrics for examining the level of property X within a system. [67] If a system scores above a certain level on that rubric, the regulator might then require that certain red teaming activities, benchmarks, and audit logs be completed before the institution can proceed with that system. Importantly, the role of expertise—both for regulators and those being regulated—is essential for detecting systems of concern. Without an adequate understanding, for example, of unwanted system behaviours and their potential dangers, it is difficult for system owners and regulators to make good use of rubrics.

Third, given the relatively large number of actors who might be subject to scrutiny, the regulator must be able to *effectively operate under conditions of uncertainty and with limited resources.* While not carried out by a single nation-state, the enforcement of nuclear safeguards by the

International Atomic Energy Agency (IAEA) provides a relevant example. Each year on a limited budget, a relatively small number of IAEA inspectors are challenged to inspect a large number of nuclear facilities worldwide; since the IAEA cannot reasonably inspect every single nuclear reactor or spent fuel rod, the agency has embraced statistical approaches centred around random sampling and remote surveillance. [68] Even if imperfect and ultimately unable to *prevent* illicit activities outright, the IAEA's system of inspections has still often succeeded in detecting violations. [69] Similarly, a regulatory body seeking to limit property X might aim to gather as much information as possible with a primary emphasis on early detection for suspicious or concerning activities; doing so might involve the use of similar methods for handling large quantities of data to ease the burden on regulators.

Finally, regulators require *enforcement capacity*, including sufficient access to system designers and operators to explore the potential presence of property X. To ensure adherence to guidelines, regulators must be granted access to relevant employees (for interviews) and facilities (for inspection) if concerns arise. A number of arrangements could result in raising concerns: While entities might be required to prove compliance to begin research and/or remain in operation, they could also be subjected to regular and/or random inspections. A more loosely crafted regulatory framework might rely on whistle-blowers to come forward, only investigating concerns if they were raised internally. Alternatively, the interactions with the regulator might be dictated by certain milestones in a system lifecycle, with regular inspections scheduled according to the speed and scope of AI research. Any of the above formulations would depend upon a regulator's ability to speak directly with system owners, potentially interviewing them about concerning behaviour displayed by the system. As a component of enforcement, regulators might also require system owners and operators to undergo mandatory trainings to increase the level of knowledge about property X and the means of avoiding it.

## 5 Conclusion

We have outlined a set of potential intrinsic characteristics of advanced AI systems that together we label "Property X". These include characteristics discussed in the AI safety literature such as power-seeking behaviour, instrumental rationality, and strategic awareness that may lead to risky behaviours such as the pursuit of convergent instrumental goals. Systems high in Property X may prove to be exceptionally capable and so may be an appealing goal of research; however, we have argued that such systems are highly likely to be dangerous and difficult to align. The research visions outlined in *Collective Intelligence, Human-Centred AI,* and *Comprehensive AI services* provide compelling alternatives that demonstrate that progress across scientific, economic and societal domains can be supported by AI systems low in Property X. While the set of characteristics that make up Property X are not yet fully specified, it is nonetheless possible to begin envisioning governance regimes that identify and address Property X behaviours in the AI development process. Our proposals are intended as a starting sketch for doing so.